\documentclass{article}
\usepackage{iclr2027_conference,times}

\usepackage{amsmath,amsfonts,bm}

\def\eqref#1{equation~\ref{#1}}

\def\1{\bm{1}}

\DeclareMathAlphabet{\mathsfit}{\encodingdefault}{\sfdefault}{m}{sl}
\SetMathAlphabet{\mathsfit}{bold}{\encodingdefault}{\sfdefault}{bx}{n}

\usepackage{booktabs}
\usepackage{float}
\usepackage{multirow}
\usepackage{graphicx}
\usepackage{subcaption}
\usepackage{xcolor}
\usepackage{pgfplots}
\pgfplotsset{compat=1.18}

\usepackage{amsmath}
\usepackage{amssymb}
\usepackage[colorlinks=true, allcolors=blue]{hyperref}
\usepackage{url}

\title{Bilinear Optimization Divergence: Diagnosing Factor-Constrained LoRA Continual Learning}
\author{YongShun Wang \quad JianLin Su \quad Yong Ma\thanks{Corresponding author.}}

\iclrfinalcopy

\begin{document}
\maketitle
\lhead{Preprint}
\hypersetup{pdftitle={Bilinear Optimization Divergence: Diagnosing Factor-Constrained LoRA Continual Learning},pdfauthor={YongShun Wang, JianLin Su, Yong Ma}}

\begin{abstract}
Orthogonality in a LoRA factor does not by itself specify what the composed update protects: the answer depends on the task-start state, the parameterization, and the realized optimizer displacement. We formalize this question through Bilinear Optimization Divergence (BOD), an anchor-relative diagnostic of effective-update response on selected historical features. The finite-step analysis distinguishes two cases. In a shared adapter, protecting the routing displacement leaves a learned-anchor residual through the changing companion factor. In a fresh zero-output block, a feasible routing state can protect the composed update while both current factors remain trainable. These conditions yield Semi-Frozen Orthogonal Routing (SFOR) for shared adapters and current-block hard protection for cumulative O-LoRA; Weight Residual Projection (WRP) enforces the required displacement after the optimizer step. Controlled two-task traces verify the predicted residual paths, reducing normalized historical response from $19.12\%$ to $0.005\%$ in the shared family and from $7.72\%$ to $0.002\%$ in the cumulative family. Four-task experiments on Qwen3-8B characterize the resulting trade-offs: SFOR improves backward transfer (BWT) from $-2.47$ to $-0.86$ with nearly unchanged average accuracy (AA), while O-LoRA hard protection improves three-order mean AA from $80.27\%$ to $81.30\%$ and forgetting measure (FM) from $2.20$ to $0.43$. Component controls also show that stricter feasibility need not improve final task performance. Together, the analysis and evidence provide an architecture-conditioned account of which constraint to enforce, how to enforce it, and how to interpret its empirical value.
\end{abstract}
\begin{center}\small Code: \url{https://github.com/legend91019/My_first}\end{center}

\section{Introduction}
\label{sec:introduction}

\begin{figure}[t]
    \centering
    \includegraphics[width=\textwidth]{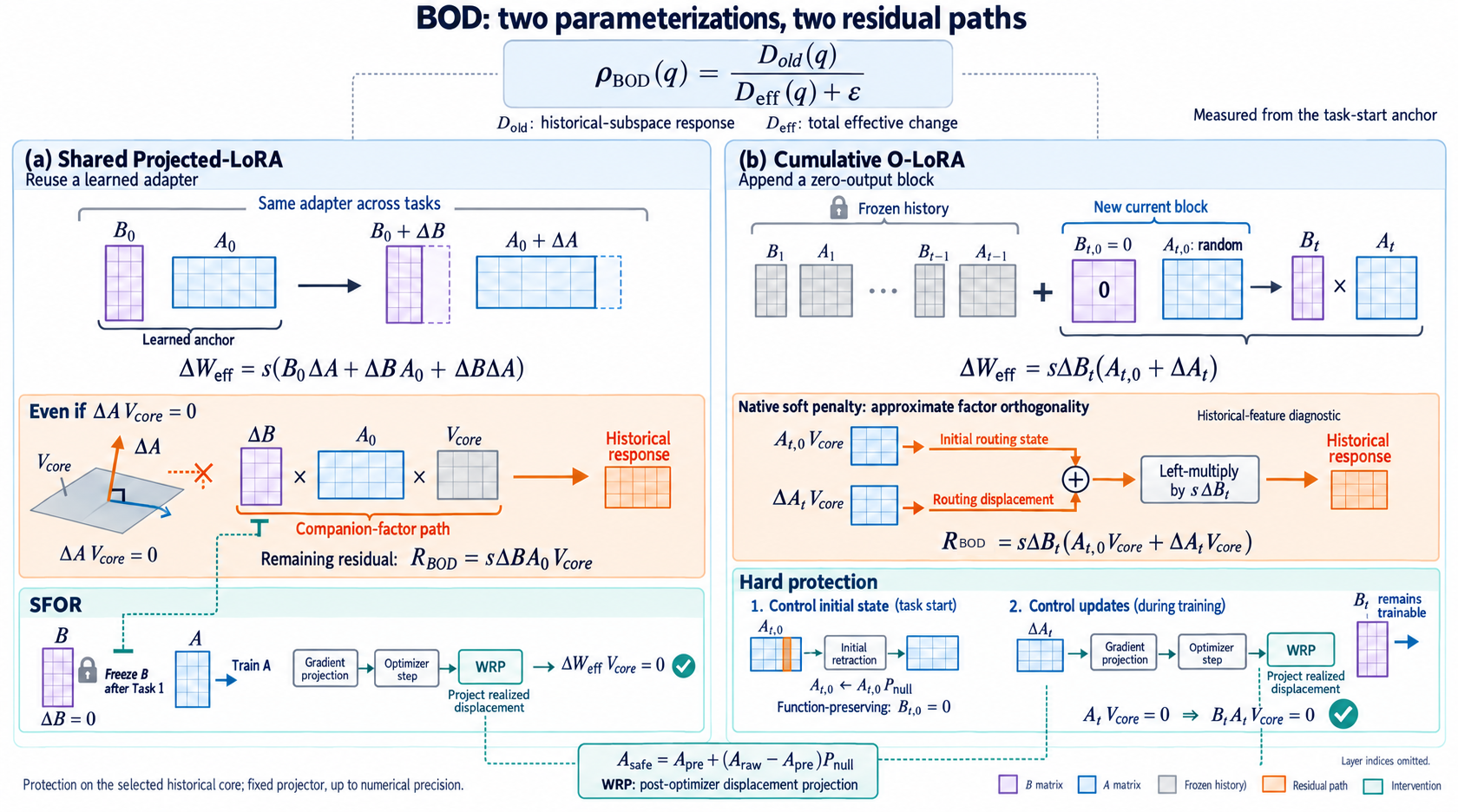}
    \caption{BOD diagnosis and architecture-specific protection. The normalized historical response is measured relative to the task-start anchor. Left: even an exactly protected routing displacement leaves a companion-factor residual in a shared adapter; SFOR freezes $B$ and constrains the realized $A$ displacement. Right: a new O-LoRA block requires protection of both its initial routing state and subsequent displacement, while its current $B$ remains trainable. WRP follows the optimizer step. $q$ indexes optimizer steps; $R_{\mathrm{BOD}}$ is a layerwise residual matrix, $D_{\mathrm{old}}$ its cross-layer root-sum-of-squares norm, and $D_{\mathrm{eff}}$ the corresponding effective-update norm. Their ratio $\rho_{\mathrm{BOD}}$ uses a positive numerical stabilizer $\varepsilon$.}
    \label{fig:bod_overview}
\end{figure}

Continual adaptation of large language models must balance acquiring new tasks with retaining earlier capabilities. LoRA~\citep{hu2021lora} makes this adaptation parameter-efficient, while orthogonality constraints seek to restrict interference. The practical question is what these restrictions actually protect: when a constraint is imposed on one LoRA factor, does the composed update satisfy the intended condition after finite optimizer steps?

A closely related family of continual-learning methods introduces orthogonality into LoRA by assigning different tasks to separated low-rank update directions. We use the convention
\begin{equation}
    \Delta W = sBA,
    \qquad
    A \in \mathbb{R}^{r\times d_{\mathrm{in}}},
    \quad
    B \in \mathbb{R}^{d_{\mathrm{out}}\times r},
\end{equation}
where $r$ is the adapter rank, $d_{\mathrm{in}}$ and $d_{\mathrm{out}}$ are layer dimensions, $s$ is the LoRA scale ($s=\alpha/r$, with scaling parameter $\alpha$), $A$ is the input-side routing factor, and $B$ is the output-side basis factor. O-LoRA~\citep{wang2023orthogonal} uses its routing factor as a proxy for historical gradient subspaces and softly encourages the current factor to be orthogonal to previous ones. InfLoRA~\citep{liang2024inflora} instead fixes a task-dependent input basis while learning its companion factor, and KeepLoRA~\citep{luo2026keeplora} combines residual-subspace routing with protection of pretrained-weight directions. Together, these methods motivate a subspace-controller view of one factor, but their guarantees are not identical. A fixed feasible state can imply an exact composed constraint, whereas soft factor orthogonality or gradient-only projection need not. We therefore ask, method by method, what effective-update condition is actually implied by the implemented factor constraint.

We examine this question with Projected-LoRA, our shared-adapter construction inspired by \emph{Sculpting Subspaces}~\citep{nayak2026sculpting}. That method constrains full-weight updates using weight singular subspaces. Our construction instead projects a shared routing gradient onto the complement of historical input features, $\nabla_A\mathcal{L}\leftarrow\nabla_A\mathcal{L}P_{\mathrm{null}}$, where $\mathcal{L}$ is the training loss, $V_{\mathrm{core}}$ contains orthonormal historical feature directions, and $P_{\mathrm{null}}=I-V_{\mathrm{core}}V_{\mathrm{core}}^\top$ is their complementary projector. Both factors remain trainable. It provides a controlled host for analyzing this change of parameterization.

Let $(A_0,B_0)$ be the task-start anchor and $(\Delta A,\Delta B)$ the subsequent factor changes. The exact effective change is $s(B_0\Delta A+\Delta B A_0+\Delta B\Delta A)$. Thus, whenever the companion factor is updated, the decomposition necessarily contains the anchor-relative path $s\Delta B A_0V_{\mathrm{core}}$; in the shared Projected-LoRA setting $A_0V_{\mathrm{core}}$ is generally nonzero, so this path can produce a nonzero response even when $\Delta A V_{\mathrm{core}}=0$. Gradient-only projection need not even ensure that displacement condition. We call this constraint-to-update gap \emph{Bilinear Optimization Divergence} (BOD) and measure it through the normalized response on protected historical directions. 

The diagnosis yields two interventions (Figure~\ref{fig:bod_overview}). Semi-Frozen Orthogonal Routing (SFOR) freezes the shared $B$ after Task~1 and applies gradient projection and post-optimizer Weight Residual Projection (WRP) to $A$. In cumulative O-LoRA, each new block for task $t$ instead starts from $B_{t,0}=0$ and random $A_{t,0}$, giving the task-local change $s\Delta B_t(A_{t,0}+\Delta A_t)$. Hard protection retracts the initial routing state and constrains its subsequent displacement while keeping the current $B_t$ trainable. We separately study effective-update regularization as a diagnostic control. Section~\ref{sec:methodology} gives the finite-step derivation and optimizer-feasibility conditions.

Our central result is a constraint-selection rule: inspect the anchor before deciding which factor to constrain. A learned shared anchor calls for control of the companion-factor path as well as the routing displacement; a zero-output new block permits a function-preserving feasible initialization. We evaluate this rule through within-family comparisons: SFOR against Projected-LoRA, and hard protection against O-LoRA. Qwen3-8B provides two levels of evidence, complemented by an Order-1 check on Mistral-7B-v0.3. Matched two-task traces test the residual paths and their suppression. Four-task experiments characterize the acquisition--retention outcomes of the resulting designs. This separates the implementation question---whether the intended geometry is realized---from the empirical question of whether that geometry benefits the task sequence.

Our contributions are summarized as follows:
\begin{itemize}
    \item An anchor-relative, finite-step formulation of BOD that distinguishes constraints on factor gradients, realized displacements, and complete states, measured by the scale-normalized historical response $\rho_{\mathrm{BOD}}$.
    \item An architecture-conditioned protection rule: SFOR removes the shared learned-anchor path, whereas cumulative hard protection maintains a feasible current routing state with trainable $B$. WRP connects both designs to actual optimizer displacements.
    \item An empirical analysis separating constraint execution from task utility: controlled traces verify the residual paths, and within-family comparisons on two backbones characterize acquisition and retention, including cases where stricter feasibility adds no performance benefit.
\end{itemize}

\section{Related Work}
\label{sec:related_work}

\subsection{Continual Learning and Subspace Protection}
\label{sec:related_continual_peft}
\label{sec:related_orthogonal_cl}

Continual learning addresses forgetting through replay~\citep{de2019episodic}, parameter or output regularization~\citep{kirkpatrick2017overcoming,li2017learning}, and task-specific modules~\citep{wang2024comprehensive}. GEM constrains episodic losses~\citep{lopez2017gradient}; OGD uses past-output gradients~\citep{farajtabar2020orthogonal}; OWM and GPM use historical input or activation information to restrict updates~\citep{zeng2019continual,saha2021gradient}. TRGP further uses task correlations to enable forward transfer~\citep{lin2022trgp}. These approaches motivate subspace protection while distinguishing retention from transfer. We ask when a factor-level constraint implies the intended constraint on a composed LoRA update.

\subsection{Orthogonal LoRA and Factor-Constrained PEFT}
\label{sec:related_orthogonal_lora}

Under $\Delta W_t=sB_tA_t$, O-LoRA freezes historical blocks and softly encourages current routing factors to be orthogonal to past ones~\citep{wang2023orthogonal}. InfLoRA fixes a routing basis constructed from residual input features~\citep{liang2024inflora}; KeepLoRA also protects pretrained-weight directions~\citep{luo2026keeplora}. Orthogonal Adaptation uses a related fixed-basis design in diffusion adaptation~\citep{po2024orthogonal}. A feasible state $AV_{\mathrm{core}}=0$ implies $BAV_{\mathrm{core}}=0$ for arbitrary $B$. These constructions illustrate why a fixed feasible state and a protected displacement from a learned anchor give different guarantees.

Freezing one LoRA factor is established: LoRA-FA freezes input-side $A$ and learns $B$ for memory-efficient adaptation~\citep{zhang2023lora}. \citet{zhu2024asymmetry} analyze the distinct roles of the two factors, while LoRA+ assigns them different learning rates to improve feature learning~\citep{hayou2024lora+}. SFOR instead freezes the learned output-side $B$ after Task~1 to remove a specific anchor-relative residual. Its motivation is historical-feature protection, with an acquisition cost evaluated separately. Other continual PEFT designs include decoupled adapters, shared attention, prompts, and selective retention~\citep{wu2025sd,he2025cl,zhao2024sapt,razdaibiedina2023progressive,qin2021lfpt5,wang2024inscl,he2024seekr}.

\subsection{Effective-Update Geometry and Positioning}
\label{sec:related_positioning}

\citet{shuttleworth2026lora} diagnose spectral differences between LoRA and full fine-tuning through intruder dimensions and study their connection to forgetting. BOD measures anchor-relative response on a selected historical feature core. Projected-LoRA is our diagnostic construction inspired by the full-weight singular-subspace constraints of \emph{Sculpting Subspaces}~\citep{nayak2026sculpting}, using historical features as in GPM~\citep{saha2021gradient}. Recent concurrent and adjacent methods address related goals from different levels: Janus-LoRA uses online subspace estimation and gradient rectification to construct safer composite factor updates~\citep{chen2026janus}; SplitLoRA partitions the historical gradient space to balance stability and plasticity~\citep{qiu2026splitlora}; CoSO continually rotates the optimization subspace using gradient SVD~\citep{cheng2026continuous}; and EBLoRA studies spectral imbalance in low-rank continual adaptation~\citep{gu2026spectral}. These methods select or reshape the protected update space, whereas our contribution concerns how a selected constraint transfers through bilinear adapters: the task anchor determines the residual path, and the realized optimizer displacement determines feasibility. SFOR and cumulative hard protection instantiate these conditions rather than introduce a new generic projection operation.

\section{Methodology}
\label{sec:methodology}

Figure~\ref{fig:bod_overview} summarizes the two parameterizations and their residual paths. We first formalize the constraint-to-update mismatch, then derive SFOR and WRP for shared adapters and hard protection for cumulative O-LoRA. All claims concern responses on selected historical-feature directions.

\subsection{Continual LoRA Setup and Historical Subspaces}
\label{sec:method_setup}

Consider a pretrained language model with weight matrices $\{W_0^{(\ell)}\}_{\ell\in\mathcal{I}}$, where $\mathcal{I}$ is the set of layers equipped with LoRA modules. The model is adapted to $N$ tasks $\{\mathcal{T}_1,\ldots,\mathcal{T}_N\}$, indexed by $t$. For a layer $\ell$, we use the convention
\begin{equation}
    \Delta W^{(\ell)}=s_\ell B^{(\ell)}A^{(\ell)},
    \qquad
    A^{(\ell)}\in\mathbb{R}^{r\times d_{\mathrm{in}}},
    \quad
    B^{(\ell)}\in\mathbb{R}^{d_{\mathrm{out}}\times r},
\end{equation}
where $s_\ell$ is the LoRA scaling coefficient. The matrix $A$ is the input-side routing factor because it acts on the layer input first, while $B$ is the output-side basis factor. The pretrained weights remain fixed throughout training.

For each layer, we collect input features from tasks that have already been learned. Let $X_{\mathrm{old}}^{(\ell)}\in\mathbb{R}^{n_{\mathrm{old}}\times d_{\mathrm{in}}}$ denote the resulting historical feature matrix, with $n_{\mathrm{old}}$ stored token-feature rows. Its singular value decomposition is
\begin{equation}
    X_{\mathrm{old}}^{(\ell)}=U^{(\ell)}\Sigma^{(\ell)}
    \left(V^{(\ell)}\right)^{\top}.
\end{equation}
Here $U^{(\ell)}$ and $V^{(\ell)}$ contain left and right singular vectors, and $\Sigma^{(\ell)}$ contains singular values. We retain $k_\ell$ right-singular directions as the historical core,
\begin{equation}
    V_{\mathrm{core}}^{(\ell)}=V^{(\ell)}_{[:,1:k_\ell]},
    \qquad
    P_{\mathrm{null}}^{(\ell)}
    =I-V_{\mathrm{core}}^{(\ell)}
      \left(V_{\mathrm{core}}^{(\ell)}\right)^{\top}.
    \label{eq:method_projector}
\end{equation}
Here $P_{\mathrm{null}}^{(\ell)}$ projects input-side directions onto the complement of the retained historical feature subspace. A routing displacement $\Delta A^{(\ell)}$ is protected at the factor level when
\begin{equation}
    \Delta A^{(\ell)}V_{\mathrm{core}}^{(\ell)}=0.
    \label{eq:factor_protection}
\end{equation}

\subsection{Bilinear Effective Updates and BOD}
\label{sec:method_bod}

The distinction between factor-level protection and effective-update protection follows directly from the bilinear LoRA parameterization. Let $(A_0,B_0)$ denote the factors at the beginning of a new task, and write their changes during training as $\Delta A$ and $\Delta B$. The change in the LoRA weight update is exactly
\begin{equation}
    \begin{aligned}
    \Delta W_{\mathrm{eff}}
    &=s\left[(B_0+\Delta B)(A_0+\Delta A)-B_0A_0\right]\\
    &=s\left(B_0\Delta A+\Delta B A_0+\Delta B\Delta A\right).
    \end{aligned}
    \label{eq:bilinear_decomposition}
\end{equation}

For one layer, define the historical response matrix
\begin{equation}
    R_{\mathrm{BOD}}=\Delta W_{\mathrm{eff}}V_{\mathrm{core}}.
    \label{eq:bod_residual}
\end{equation}
The matrix $R_{\mathrm{BOD}}$ records layerwise responses along the retained directions. Its vanishing certifies this selected-feature constraint; task accuracy additionally depends on untracked directions and changes elsewhere in the network. Write $R_{\mathrm{BOD}}^{(\ell)}(q)$ for this matrix at layer $\ell$ and optimizer step $q$. Over the tracked layers $\mathcal{I}$, define
\begin{equation}
\begin{split}
 D_{\mathrm{old}}(q)&=\left(\sum_{\ell\in\mathcal{I}}\|R_{\mathrm{BOD}}^{(\ell)}(q)\|_F^2\right)^{1/2},\\
 D_{\mathrm{eff}}(q)&=\left(\sum_{\ell\in\mathcal{I}}\|\Delta W_{\mathrm{eff}}^{(\ell)}(q)\|_F^2\right)^{1/2},\qquad
 \rho_{\mathrm{BOD}}(q)=\frac{D_{\mathrm{old}}(q)}{D_{\mathrm{eff}}(q)+\varepsilon}.
\end{split}
\label{eq:diagnostic_drift}
\end{equation}
The Frobenius norm is denoted by $\|\cdot\|_F$, and $\varepsilon>0$ stabilizes the denominator. Thus $D_{\mathrm{old}}$ is an absolute residual magnitude, while $\rho_{\mathrm{BOD}}$ is its scale-normalized ratio, reported as $100\rho_{\mathrm{BOD}}\%$. Under exact routing-displacement protection, $\Delta A V_{\mathrm{core}}=0$, the layerwise response reduces to
\begin{equation}
 R_{\mathrm{BOD}}=s\Delta B A_0V_{\mathrm{core}}.
 \label{eq:bod_residual_reduced}
\end{equation}
Gradient-only projection can leave additional terms when the realized displacement violates this condition. For O-LoRA, the same measurement evaluates the added historical-feature constraint; its native penalty acts on historical routing factors.

The anchor determines which invariant is available. A displacement constraint $\Delta A V_{\mathrm{core}}=0$ only states that the new routing movement has no component on the selected core; it does not remove an existing anchor component $A_0V_{\mathrm{core}}$. By contrast, a state constraint $AV_{\mathrm{core}}=0$, with $A=A_0+\Delta A$, gives $BAV_{\mathrm{core}}=0$ for any current $B$. This is why the two parameterizations require different interventions: a shared learned anchor requires control of the companion-factor path, whereas a fresh zero-output block can first be initialized in the feasible state and then keep its current $B$ trainable. BOD measures selected-core response; effective drift measures update magnitude, while continual-learning metrics assess task-level outcomes.

\subsection{Projected-LoRA}
\label{sec:method_projected_lora}

Projected-LoRA transfers the subspace-projection idea of full-parameter continual fine-tuning to a shared LoRA branch. During the first task, both $A^{(\ell)}$ and $B^{(\ell)}$ are trainable. After the historical feature basis in Eq.~(\ref{eq:method_projector}) has been constructed, subsequent-task optimization projects the gradient of the shared routing factor:
\begin{equation}
    \nabla_{A^{(\ell)}}\mathcal{L}
    \leftarrow
    \nabla_{A^{(\ell)}}\mathcal{L}
    P_{\mathrm{null}}^{(\ell)}.
    \label{eq:projected_lora_gradient}
\end{equation}
Both factors remain trainable, and the number of LoRA blocks does not grow with the number of tasks. Projected-LoRA therefore implements factor-level historical routing protection while retaining the companion-factor pathway in Eq.~(\ref{eq:bilinear_decomposition}). It serves as the shared-factor host on which the effect of BOD and the SFOR interventions are isolated.

\subsection{SFOR and Weight Residual Projection}
\label{sec:method_sfor_wrp}

SFOR is derived from Projected-LoRA by structurally removing the shared companion-factor pathway after the first task. Once task $\mathcal{T}_1$ is completed, the output-side basis is frozen:
\begin{equation}
    \Delta B^{(\ell)}_t=0,\qquad t\geq 2,
    \quad \ell\in\mathcal{I}.
    \label{eq:sfor_freeze}
\end{equation}
The routing factor remains trainable and its gradient is projected according to Eq.~(\ref{eq:projected_lora_gradient}). Since $B$ is fixed, the effective update for subsequent tasks has the linear form $sB_{\mathrm{fixed}}\Delta A$ relative to the task-start anchor, eliminating the $\Delta B A_0$ and $\Delta B\Delta A$ pathways. The name ``semi-frozen'' refers to the LoRA branch as a whole: $B$ is fully frozen after task 1, whereas $A$ continues to adapt.

Gradient projection alone does not guarantee a feasible parameter displacement: Adam applies coordinate-wise preconditioning~\citep{kingma2014adam}, and AdamW additionally applies decoupled weight decay~\citep{loshchilov2017decoupled}. Appendix~\ref{app:optimizer_geometry} gives a concrete example. Let $A_{\mathrm{pre}}$ be the routing factor immediately before an optimizer step and $A_{\mathrm{raw}}$ its value immediately after the unmodified optimizer transformation. WRP projects the realized residual rather than only the gradient:
\begin{equation}
    A_{\mathrm{safe}}
    =A_{\mathrm{pre}}
     +\left(A_{\mathrm{raw}}-A_{\mathrm{pre}}\right)P_{\mathrm{null}}.
    \label{eq:wrp}
\end{equation}
Geometrically, WRP is the Euclidean projection onto the affine set $\{A:(A-A_{\mathrm{pre}})V_{\mathrm{core}}=0\}$. Its role here is to enforce the derived invariant after optimizer transformations. When the task-relative displacement accumulated before the step already satisfies Eq.~(\ref{eq:factor_protection}), Eq.~(\ref{eq:wrp}) preserves the feasible component of the optimizer step and removes its protected-subspace component. WRP is applied after every optimizer step to the trainable routing rows. SFOR therefore combines a structural constraint, $\Delta B=0$, with an explicit post-step enforcement of the routing constraint, while keeping a fixed number of LoRA parameters.

\subsection{Cumulative O-LoRA Blocks and Hard Protection}
\label{sec:method_olora_bsr}

O-LoRA uses cumulative task-specific blocks, indexed by task $t$. At layer $\ell$, the factors after task $t$ can be written as
\begin{equation}
    A^{(\ell)}=
    \begin{bmatrix}A^{(\ell)}_{<t}\\A^{(\ell)}_t\end{bmatrix},
    \qquad
    B^{(\ell)}=
    \begin{bmatrix}B^{(\ell)}_{<t}&B^{(\ell)}_t\end{bmatrix},
    \label{eq:olora_blocks}
\end{equation}
where historical blocks remain fixed and only the current block is optimized. We follow the official implementation and use its element-wise absolute-sum penalty:
\begin{equation}
    \mathcal{L}_{\mathrm{orth}}^{(t)}
    =\sum_{\ell\in\mathcal{I}}
      \left\|A_{<t}^{(\ell)}
      A_t^{(\ell)\top}\right\|_1,
    \label{eq:olora_orth}
\end{equation}
where $\|\cdot\|_1$ is the entrywise absolute sum. The O-LoRA paper and official code differ in factor naming and penalty form; here ``O-LoRA'' follows the code, with $\Delta W=BA$ and input-side factor $A$ constrained by Eq.~(\ref{eq:olora_orth}).

Each new block starts from $B_{t,0}^{(\ell)}=0$ and random $A_{t,0}^{(\ell)}$. Writing $A_t=A_{t,0}+\Delta A_t$ and $B_t=\Delta B_t$, the response of the newly learned block on the historical core is
\begin{equation}
    R_{\mathrm{BOD}}^{(\ell,t)}
    =s_\ell\Delta B_t^{(\ell)}
    \left(A_{t,0}^{(\ell)}V_{\mathrm{core}}^{(\ell)}
    +\Delta A_t^{(\ell)}V_{\mathrm{core}}^{(\ell)}\right).
    \label{eq:olora_block_residual}
\end{equation}
Unlike a shared adapter with a learned anchor, each new block starts with zero effective update. Its residual depends on both the initial routing state and subsequent displacement. O-LoRA's soft penalty encourages approximate factor orthogonality; it does not enforce zero response on the historical feature basis used here.

We evaluate hard protection as an extension of existing O-LoRA: the native training objective is retained, and protection operations are added only to the current block. Current-block hard protection addresses the two residual terms directly. At task start, it retracts only the new routing state,
\begin{equation}
    A_{t,0}^{(\ell)}\leftarrow
    A_{t,0}^{(\ell)}P_{\mathrm{null}}^{(\ell,t-1)}.
    \label{eq:olora_initial_retraction}
\end{equation}
This operation is function-preserving because $B_{t,0}=0$ and it never changes a historical block. During training, the current $A_t$ gradient is projected as
\begin{equation}
    \nabla_{A_t^{(\ell)}}\mathcal{L}
    \leftarrow
    \nabla_{A_t^{(\ell)}}\mathcal{L}
    P_{\mathrm{null}}^{(\ell,t-1)},
    \label{eq:olora_gradient_projection}
\end{equation}
and WRP projects every realized post-optimizer displacement using Eq.~(\ref{eq:wrp}). With a fixed projector, initialization retraction gives $A_{t,0}V_{\mathrm{core}}=0$ and WRP maintains $\Delta A_tV_{\mathrm{core}}=0$; Eq.~(\ref{eq:olora_block_residual}) then vanishes for any learned $\Delta B_t$. Thus, feasible initialization plus strict $A_t$-displacement protection suffices without freezing $B_t$. In a shared adapter, displacement protection alone leaves $s\Delta BA_0V_{\mathrm{core}}$. Hard protection maintains the routing-state invariant while the current $B_t$ learns freely, and is therefore our primary O-LoRA extension. Its empirical evaluation concerns the complete extension of O-LoRA.

\paragraph{Diagnostic effective-update extension.}
We additionally test Bilinear Stabilization Regularization (BSR) as a diagnostic control: it penalizes the current block's effective change over a wider historical basis. BSR complements the hard-protection analysis but is not required for the selected-core feasibility guarantee. Its objective is specified in Appendix~\ref{app:bsr_objective}, and its performance is compared with hard protection in Appendix Table~\ref{tab:olora_bsr_ablation}.

\label{sec:method_diagnostics}
\section{Experiments}
\label{sec:experiments}

\subsection{Experimental Setup}
\label{sec:experimental_setup}

\subsubsection{Models and Datasets}

We conduct the main experiments with Qwen3-8B~\citep{yang2025qwen3}. Following the operational Standard CL protocol used by O-LoRA~\citep{wang2023orthogonal}, N-LoRA~\citep{yang2025parameter}, and LFPT5~\citep{qin2021lfpt5}, the primary evaluation uses four text-classification tasks under three task orders. All main runs use seeds $\{38,42,2026\}$. Dataset details, task sequences, and hyperparameters are provided in the appendix.

\subsubsection{Baselines}

We compare against sequential and cumulative LoRA, replay, parameter/output regularization, and orthogonal-routing baselines. Specifically, the suite contains Seq-LoRA, IncLoRA, Replay, EWC~\citep{kirkpatrick2017overcoming}, LwF~\citep{li2017learning}, O-LoRA~\citep{wang2023orthogonal}, N-LoRA~\citep{yang2025parameter}, SD-LoRA~\citep{wu2025sd}, and our Projected-LoRA construction (Section~\ref{sec:related_positioning}). Unless a method-specific design requires otherwise, all methods use the same backbone, target modules, LoRA rank, optimizer, and training protocol. N-LoRA uses the causal-LM objective $\mathcal{L}_{\mathrm{CE}}+\lambda\lVert sB_tA_t\rVert_1$; SD-LoRA uses one normalized block per task with magnitude gates; Projected-LoRA projects the shared $A$-gradient onto the historical null space while keeping both factors trainable. Full implementation details are given in Appendix~\ref{app:protocol}.

\subsubsection{Metrics}

We report Average Accuracy (AA), Learning Accuracy (LA), Backward Transfer (BWT), and the signed Forgetting Measure (FM). AA is the final average performance over all tasks. LA measures immediate acquisition of each task. BWT measures the average change in previously learned task performance after subsequent-task training, with values closer to zero indicating less forgetting. FM measures the average change from the best pre-final checkpoint to the final checkpoint; lower values indicate better retention, and negative values indicate positive transfer beyond the previous best.

Mechanism traces report $D_{\mathrm{eff}}$, $D_{\mathrm{old}}$, and $\rho_{\mathrm{BOD}}$ from Eq.~(\ref{eq:diagnostic_drift}). Additional drift definitions and the continual-learning metric formulas are collected in Appendix~\ref{app:metrics}.

\subsection{Two-Task Mechanism Evidence}
\label{sec:mechanism_tracking}

\begin{figure}[t]
\centering
\includegraphics[width=\textwidth]{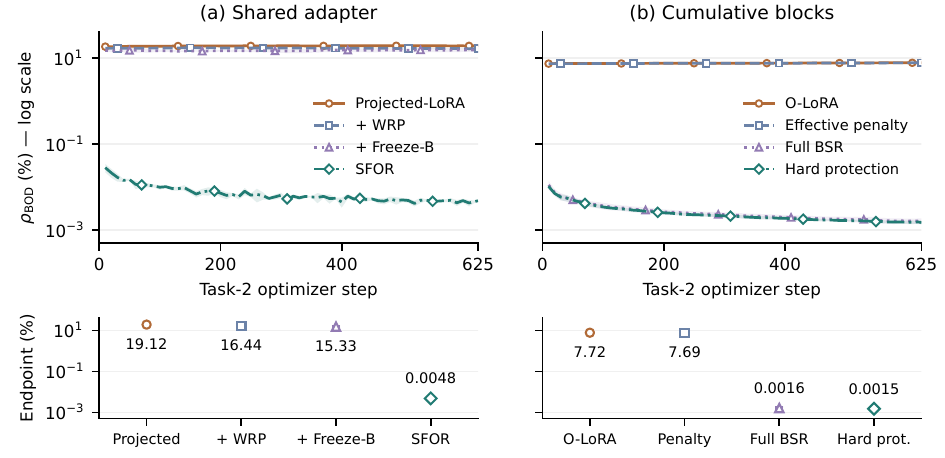}
\caption{Microscopic BOD dynamics on DBpedia $\rightarrow$ Amazon. Top: trajectories with distinct markers at staggered recorded steps. Bottom: step-625 means and one seed standard deviation ($n=3$), with numerical percentages. Bands also show one seed standard deviation. Both rows use the same logarithmic scale; step zero is omitted. Coincident curves indicate similar measured responses. Shared-adapter controls separate freezing $B$ from enforcing its companion's displacement. For cumulative blocks, hard protection suppresses the residual while the effective penalty alone does not. }
\label{fig:main_bod_dynamics}
\end{figure}

We use a controlled DBpedia $\rightarrow$ Amazon case study on Qwen3-8B to examine effective-update drift and historical-subspace response. The trace compares the Projected-LoRA and O-LoRA component families over 625 Task-2 optimizer steps and 64 recorded points. At the endpoint, $\rho_{\mathrm{BOD}}$ is $0.005\%$ for SFOR and $0.002\%$ for O-LoRA hard protection, compared with $19.12\%$ for Projected-LoRA and $7.72\%$ for vanilla O-LoRA. These microscopic measurements test the residual paths and their suppression on the selected historical subspace. Figure~\ref{fig:main_bod_dynamics} shows the normalized trajectories. Additional diagnostics, seed variability, and probe details are in Appendices~\ref{app:mechanism_results} and~\ref{app:uncertainty}.

The controls distinguish two paths: WRP alone leaves a $16.44\%$ residual in the shared family, while Freeze-$B$ alone leaves $15.33\%$; their combination reaches $0.005\%$. This is consistent with jointly controlling the companion factor and optimizer displacement. Exact endpoints appear in Appendix Tables~\ref{tab:sfor_family_mechanism} and~\ref{tab:olora_family_mechanism}.

\subsection{Four-Task Acquisition and Retention}
\label{sec:main_results}

Table~\ref{tab:comprehensive_qwen8b} reports the three-order benchmark. SFOR leaves mean AA nearly unchanged ($80.38\%$ versus $80.45\%$), improves BWT from $-2.47$ to $-0.86$, and FM from $2.50$ to $0.91$, with lower immediate acquisition (Appendix Table~\ref{tab:complete_cl_metrics}). O-LoRA hard protection improves mean AA from $80.27\%$ to $81.30\%$ and FM from $2.20$ to $0.43$, also with a small acquisition decrease. The shared design primarily shifts the balance toward retention at fixed adapter size; the cumulative design also improves final mean accuracy. 

\begin{table*}[t]
\centering
\small
\caption{Standard CL results with Qwen3-8B. Each row averages the three task orders and seeds $\{38,42,2026\}$. Full metrics including LA appear in Appendix Table~\ref{tab:complete_cl_metrics}; seed statistics are in Appendix~\ref{app:uncertainty}. AA is a percentage; BWT and signed FM are percentage points. Bold and underline denote the best and second-best values.}
\label{tab:comprehensive_qwen8b}
\begin{tabular}{lccc}
\toprule
\textbf{Method} & \textbf{AA} & \textbf{BWT} & \textbf{FM} \\
\midrule
\multicolumn{4}{l}{\emph{Shared adapter: fixed stored rank}} \\
Projected-LoRA & 80.38 & -2.47 & 2.50 \\
SFOR & 80.45 & \underline{-0.86} & \underline{0.91} \\
\midrule
\multicolumn{4}{l}{\emph{Cumulative adapter: matched block growth}} \\
O-LoRA & 80.27 & -2.13 & 2.20 \\
O-LoRA + hard protection & \textbf{81.30} & \textbf{-0.38} & \textbf{0.43} \\
\midrule
\multicolumn{4}{l}{\emph{Other rehearsal-free references}} \\
EWC & 80.87 & -1.56 & 1.59 \\
IncLoRA & 80.24 & -2.43 & 2.49 \\
SD-LoRA & 80.19 & -2.48 & 2.53 \\
N-LoRA & 80.12 & -2.21 & 2.28 \\
LwF & 79.90 & -1.50 & 1.53 \\
Seq-LoRA & 79.68 & -3.42 & 3.47 \\
\midrule
\multicolumn{4}{l}{\emph{Replay reference: historical examples available}} \\
Replay & \underline{81.13} & -1.30 & 1.39 \\
\bottomrule
\end{tabular}
\end{table*}

\subsection{Cross-backbone check}
\label{sec:cross_backbone}

To check whether the main comparison depends on a single backbone, we repeat the five-method comparison on a Mistral-7B-v0.3 using the same four-task Order-1 protocol and seeds $\{38,42,2026\}$. The table reports means over the three seeds; seed-level standard deviations are reported in Appendix~\ref{app:uncertainty}. This experiment is a backbone transfer check, while the BOD mechanism analysis remains anchored to Qwen3-8B.

\begin{table}[t]
\centering
\small
\caption{Order-1 cross-backbone results on the Mistral-7B-v0.3. AA is higher-is-better; BWT and FM are reported in percentage points, with FM lower-is-better.}
\label{tab:cross_backbone}
\begin{tabular}{lccc}
\toprule
\textbf{Method} & \textbf{AA} & \textbf{BWT} & \textbf{FM} \\
\midrule
Seq-LoRA & 70.42 & $-15.12$ & $15.19$ \\
O-LoRA & 79.30 & $-4.62$ & $4.66$ \\
O-LoRA + hard protection & 80.99 & $-2.45$ & $2.47$ \\
Projected-LoRA & 75.04 & $-9.77$ & $9.79$ \\
SFOR & \textbf{81.67} & \textbf{$-1.13$} & \textbf{$1.17$} \\
\bottomrule
\end{tabular}
\end{table}

Both within-family comparisons improve in this Order-1 cohort: SFOR raises mean AA by $6.63$ percentage points over Projected-LoRA, and hard protection raises it by $1.69$ over O-LoRA. AA, BWT, and FM improve for each matched seed in both comparisons (Appendix Table~\ref{tab:mistral_paired}). Mean learning accuracy changes by $+0.15$ and $+0.07$ points, respectively. Thus, the observed retention gains here do not accompany a decrease in mean acquisition; they complement the acquisition--retention trade-off seen for SFOR on Qwen3-8B.

\subsection{Ablation Studies}
\label{sec:ablation}

\begin{figure}[t]
\centering
\includegraphics[width=\textwidth]{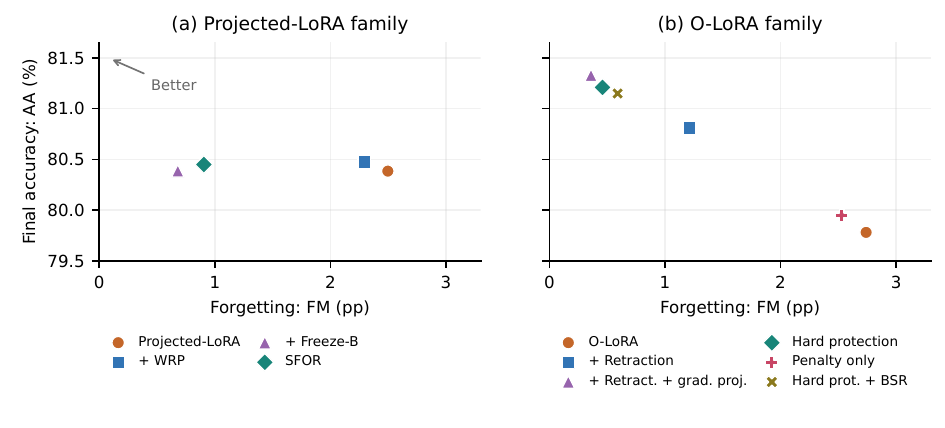}
\caption{AA--FM outcomes; upper left is better. Colors and marker shapes identify variants within each family. The two panels summarize the shared-adapter and cumulative-block component cohorts, respectively. The shared panel shows the complementarity of Freeze-$B$ and WRP, while the cumulative panel shows the hard-protection trade-off. Full metrics appear in Appendix~\ref{app:ablation_details}.}
\label{fig:main_olora_ablation}
\end{figure}

The shared family separates WRP, Freeze-$B$, and their combination in the matched mechanism trace. As shown in Figure~\ref{fig:main_olora_ablation} and Appendix Table~\ref{tab:projected_component_ablation}, either component alone leaves a substantial historical response, whereas their combination gives the strongest BOD suppression and the best retention trade-off for the shared adapter. For matched Order-1 O-LoRA, initialization retraction and gradient projection progressively improve FM, and the complete hard-protection configuration reports the corresponding cumulative result (Appendix Table~\ref{tab:olora_bsr_ablation}). The effective-update regularizer is retained as a diagnostic comparison; its full metrics are reported in Appendix~\ref{app:ablation_details}.

\paragraph{What the diagnosis changes.}
BOD supports a staged decision: first identify whether the intended invariant concerns a state or a displacement; then enforce the corresponding finite-step condition; finally assess acquisition and retention together. In the shared adapter, WRP and Freeze-$B$ act as complementary parts of SFOR: their combination suppresses the residual path and improves the observed acquisition--retention trade-off. In cumulative O-LoRA, hard protection provides the complete state-and-displacement construction, while BSR serves as an additional effective-update control.

\paragraph{Scope.}
The mechanism probe and benchmark study use Qwen3-8B; the additional Mistral-7B-v0.3 check provides a cross-backbone Order-1 comparison. Guarantees apply to the stored feature core, whose sampling and rank limits are specified in Appendix~\ref{app:experimental_details}. The experiments characterize the observed acquisition--retention trade-offs and the feasibility of the proposed constraints. Seed SD describes variability, while the paired Mistral comparisons report matched per-seed differences.

\section{Conclusion}
\label{sec:conclusion}

BOD provides a finite-step account of how factor constraints translate into composed-update protection. Its central distinction is architectural: a shared learned anchor leaves a companion-factor path, whereas a fresh zero-output block permits feasible-state initialization. SFOR and cumulative hard protection implement the corresponding conditions, with WRP enforcing realized displacements. The traces verify these mechanisms, and the four-task results characterize their retention benefits and acquisition costs. The resulting practical lesson is to choose the invariant from the parameterization, verify it on effective updates, and evaluate its task-level utility as a separate question.

\clearpage
\bibliographystyle{iclr2027_conference}
\bibliography{sample}

\subsection*{AI Use Statement}

We used generative AI tools to assist with language editing, paper organization, and LaTeX formatting. Generative AI also assisted with figure design and plotting code based on existing results. AI assistance also supported experiment scripting, execution, and result auditing. Experimental measurements were obtained from model runs; AI-generated estimates were not used as data. All AI-assisted text and code suggestions were reviewed and verified by the authors. We take responsibility for the final content of this work, including all claims, experiments, and artifacts produced with the aid of generative AI.

\subsection*{Reproducibility Statement}

The paper specifies the parameterization and update rules for Projected-LoRA, SFOR, Weight Residual Projection, O-LoRA hard protection, and the diagnostic BSR extension, together with the continual-learning and mechanism-level metrics. The experimental protocol, benchmark descriptions, projector construction, selection gate, and additional derivations are provided in the appendix. The implementation and reproduction instructions are available at \url{https://github.com/legend91019/My_first}.

\clearpage
\appendix
\section{Derivations}

\label{app:bod_derivations}

This appendix derives the two residual paths and the invariant maintained by WRP. Symbols follow the main text throughout.

\subsection{Exact Bilinear Identity and No-Leakage Condition}
\label{app:bilinear_identity}

For one LoRA layer, let $W=W_0+sBA$ and let $(A_0,B_0)$ be an anchor. The exact effective update relative to the anchor is
\begin{equation}
    \Delta W_{\mathrm{eff}}=s(BA-B_0A_0).
    \label{eq:app_exact_update}
\end{equation}
Writing $A=A_0+\Delta A$ and $B=B_0+\Delta B$ yields
\begin{equation}
    \Delta W_{\mathrm{eff}}
    =s\left(B_0\Delta A+\Delta B A_0+\Delta B\Delta A\right).
    \label{eq:app_bilinear_identity}
\end{equation}
Let $V_{\mathrm{core}}$ contain retained historical right-singular directions and $P_{\mathrm{null}}=I-V_{\mathrm{core}}V_{\mathrm{core}}^{\top}$. If the stronger invariant $A=AP_{\mathrm{null}}$ is enforced exactly at every step, then $AV_{\mathrm{core}}=0$ and
\begin{equation}
    BAV_{\mathrm{core}}=0
    \qquad\text{for any compatible }B.
    \label{eq:app_exact_noleakage}
\end{equation}
For an anchor-relative displacement, the weaker condition $\Delta A V_{\mathrm{core}}=0$ gives
\begin{equation}
    \Delta W_{\mathrm{eff}} V_{\mathrm{core}}
    =s\,\Delta B A_0 V_{\mathrm{core}},
    \label{eq:app_anchor_leakage}
\end{equation}
because the terms containing $\Delta A V_{\mathrm{core}}$ vanish. Hence, if $A_0V_{\mathrm{core}}\neq 0$, a trainable companion factor can still produce an effective residual on the retained directions. Conversely, if the complete state satisfies $AV_{\mathrm{core}}=0$ (for example, after a feasible initialization followed by exact WRP), then Eq.~(\ref{eq:app_exact_noleakage}) holds for any compatible $B$. These two cases distinguish a factor-level displacement constraint from a stronger state-level invariant.

At the end of task $t$, the operational historical residual for layer $\ell$ is
\begin{equation}
    R_{\mathrm{BOD}}^{(\ell,t)}=\Delta W_{\mathrm{eff}}^{(\ell,t)}V_{\mathrm{core}}^{(\ell)}.
    \label{eq:app_residual}
\end{equation}
A non-negligible $\|R_{\mathrm{BOD}}^{(\ell,t)}\|_F$ indicates that the implemented protection does not enforce the corresponding effective-update constraint on the retained directions. Because $R_{\mathrm{BOD}}^{(\ell,t)}$ depends on the product $BA$, this residual is invariant to the factor rescaling $BA=(cB)(A/c)$ for any nonzero scalar $c$.

\subsection{BOD for Cumulative O-LoRA Blocks}
\label{app:olora_bod}

For cumulative task blocks, the layer update after task $t$ is
\begin{equation}
    W_t^{(\ell)}
    =W_0^{(\ell)}+s_\ell\sum_{j=1}^{t}B_j^{(\ell)}A_j^{(\ell)}.
    \label{eq:app_cumulative_update}
\end{equation}
Historical blocks are frozen, while the current block $(A_t,B_t)$ remains trainable. Following the official implementation, the O-LoRA penalty controls the factor-level relation
\begin{equation}
    \left\|A_{<t}^{(\ell)}(A_t^{(\ell)})^{\top}\right\|_1,
    \label{eq:app_olora_factor_constraint}
\end{equation}
but this relation does not imply
\begin{equation}
    \left(B_t^{(\ell)}A_t^{(\ell)}-B_{t,0}^{(\ell)}A_{t,0}^{(\ell)}\right)V_{\mathrm{core}}^{(\ell)}=0.
    \label{eq:app_olora_effective_constraint}
\end{equation}
The soft penalty need not reach zero. Even exact factor orthogonality would protect the row space of $A_{<t}$, which need not contain $V_{\mathrm{core}}^{(\ell)}$. Thus, $R_{\mathrm{BOD}}^{(\ell,t)}$ is a common historical-feature response diagnostic, not a measurement of the native penalty itself. Hard protection adds an explicit constraint on $V_{\mathrm{core}}^{(\ell)}$.

The residual paths differ because the two parameterizations expose different
state histories. For a shared branch, if the input displacement is protected
so that $\Delta A_t V_{\mathrm{core}}^{(\ell)}=0$, the bilinear identity leaves the companion-factor
path
\begin{equation}
    R_{\mathrm{BOD}}^{(\ell,t)}
    =s_\ell\,\Delta B_t A_{t,0} V_{\mathrm{core}}^{(\ell)},
    \label{eq:app_shared_bod_path}
\end{equation}
where $A_{t,0}$ is the shared routing state at the beginning of task $t$. The
factor $B$ is shared and repeatedly updated, so this path can accumulate over
tasks. In cumulative O-LoRA, the new block is initialized with $B_{t,0}=0$
and random $A_{t,0}$, while all historical blocks are frozen. Writing
$B_t=\Delta B_t$ and $A_t=A_{t,0}+\Delta A_t$, its current-block residual at the end of task $t$ is
\begin{equation}
    R_{\mathrm{BOD}}^{(\ell,t)}
    =s_\ell\Delta B_t
    \left(A_{t,0}V_{\mathrm{core}}^{(\ell)}+\Delta A_tV_{\mathrm{core}}^{(\ell)}\right),
    \label{eq:app_block_bod_path}
\end{equation}
which is local to the newly allocated block. Task-start retraction makes
$A_{t,0}V_{\mathrm{core}}^{(\ell)}=0$ without changing the initialized function, and gradient
projection plus WRP maintain $\Delta A_tV_{\mathrm{core}}^{(\ell)}=0$ for a fixed projector. These
conditions eliminate the measured current-block residual for arbitrary
$\Delta B_t$. Vanilla O-LoRA's native factor-orthogonality relation does not
imply either condition, but task-wise allocation still avoids the same
cross-task accumulation mechanism as a shared-factor method.

\subsection{WRP and Optimizer Feasibility}
\label{app:wrp}
\label{app:wrp_identity}

Let $P_{\mathrm{null}}=I-V_{\mathrm{core}}V_{\mathrm{core}}^{\top}$ be fixed during one task, and let $A_{\mathrm{anchor}}$ be the task-start routing factor. Suppose that the accumulated displacement before the step satisfies $(A_{\mathrm{pre}}-A_{\mathrm{anchor}})V_{\mathrm{core}}=0$. Let $A_{\mathrm{raw}}$ be the state produced by the optimizer before post-step correction. WRP sets
\begin{equation}
    A_{\mathrm{safe}}
    =A_{\mathrm{pre}}
    +(A_{\mathrm{raw}}-A_{\mathrm{pre}})P_{\mathrm{null}}.
    \label{eq:app_wrp}
\end{equation}
Since $P_{\mathrm{null}}V_{\mathrm{core}}=0$, the realized step after WRP has no component on the retained directions:
\begin{equation}
    (A_{\mathrm{safe}}-A_{\mathrm{pre}})V_{\mathrm{core}}
    =(A_{\mathrm{raw}}-A_{\mathrm{pre}})P_{\mathrm{null}}V_{\mathrm{core}}
    =0.
    \label{eq:app_wrp_step_invariant}
\end{equation}
The task-relative invariant is then preserved:
\begin{equation}
    (A_{\mathrm{safe}}-A_{\mathrm{anchor}})V_{\mathrm{core}}
    =(A_{\mathrm{pre}}-A_{\mathrm{anchor}})V_{\mathrm{core}}
    +(A_{\mathrm{raw}}-A_{\mathrm{pre}})P_{\mathrm{null}}V_{\mathrm{core}}
    =0.
    \label{eq:app_wrp_invariant}
\end{equation}
Thus, WRP is a projection of the optimizer residual. It is equivalent to projecting the full state only in the special case $A_{\mathrm{pre}}P_{\mathrm{null}}=A_{\mathrm{pre}}$; the residual formulation remains valid without that stronger assumption.

\paragraph{Feasible states and displacements.}
\label{app:wrp_limits}

For fixed $V_{\mathrm{core}}$, the feasible state set
\begin{equation}
    \mathcal{F}=\{A\in\mathbb{R}^{r\times d_{\mathrm{in}}}:AV_{\mathrm{core}}=0\}
    \label{eq:app_feasible_set}
\end{equation}
is a linear subspace. WRP preserves $(A_q-A_0)V_{\mathrm{core}}=0$ by induction from the zero initial displacement. If $A_0V_{\mathrm{core}}=0$ as well, then $A_qV_{\mathrm{core}}=0$ and $B_qA_qV_{\mathrm{core}}=0$ for any $B_q$. For a learned shared anchor, the remaining condition is $\Delta BA_0V_{\mathrm{core}}=0$; freezing $B$ is a sufficient implementation of this condition.

\paragraph{Optimizer transformations.}
\label{app:optimizer_question}

\label{app:optimizer_geometry}
For a fixed projector, momentum formed entirely from feasible gradients remains feasible if initialized feasibly. Coordinate-wise preconditioning need not preserve that subspace. For example, let a flattened routing row have protected direction $v=(1,1)^\top/\sqrt{2}$ and projected gradient $g=(1,-1)$, so $gv=0$. A diagonal preconditioner $H=\operatorname{diag}(1,2)$ gives $(gH)v=-1/\sqrt{2}\neq0$. Adam's coordinate-wise scaling~\citep{kingma2014adam} can have this form when its second-moment state is nonuniform; the resulting displacement can leave the protected subspace. Momentum carried across a change of projector can also be infeasible. For learning rate $\eta$ and decay coefficient $\lambda$, AdamW's decay displacement $-\eta\lambda A$ has protected response $-\eta\lambda AV_{\mathrm{core}}$~\citep{loshchilov2017decoupled}, which vanishes for a feasible state but need not for a learned shared anchor. WRP removes the protected component of the complete realized step.

\clearpage
\section{Experimental Protocol and Metrics}

\label{app:experimental_details}

\subsection{Models, Benchmarks, and Task Orders}
\label{app:datasets}

The main benchmark uses Qwen3-8B. The operational Standard CL benchmark contains four text-classification tasks evaluated under three task orders. The training sets contain 10,000 DBpedia, 5,000 Amazon, 10,000 Yahoo, and 4,000 AG News examples; evaluation uses all 7,600 entries per task in the O-LoRA CL\_Benchmark test files, not the full test splits of the original datasets. Yelp appears in the source benchmark catalog but is not part of these three four-task orders. The general-capability evaluation uses zero-shot MMLU on the final checkpoint after Order-1, without training or hyperparameter tuning on MMLU. The mechanism trace uses the DBpedia $\rightarrow$ Amazon sequence on Qwen3-8B, records 64 points over 625 Task-2 optimizer steps.

\begin{table}[H]
\centering
\small
\caption{Four-task Standard CL orders used in all main experiments.}
\label{tab:task_orders}
\begin{tabular}{ll}
\toprule
\textbf{Order} & \textbf{Task sequence} \\
\midrule
Order-1 & DBpedia $\rightarrow$ Amazon $\rightarrow$ Yahoo $\rightarrow$ AG News \\
Order-2 & DBpedia $\rightarrow$ Amazon $\rightarrow$ AG News $\rightarrow$ Yahoo \\
Order-3 & Yahoo $\rightarrow$ Amazon $\rightarrow$ AG News $\rightarrow$ DBpedia \\
\bottomrule
\end{tabular}
\end{table}

\begin{table}[H]
\centering
\small
\caption{Experiment settings explicitly specified in the current study.}
\label{tab:app_protocol_summary}
\begin{tabular}{p{0.30\linewidth}p{0.63\linewidth}}
\toprule
\textbf{Item} & \textbf{Setting} \\
\midrule
Main backbone & Qwen3-8B; cross-backbone check: Mistral-7B-v0.3 \\
Standard benchmark & Four text-classification tasks \\
Task orders & Three orders listed in Table~\ref{tab:task_orders} \\
Random seeds & $38,42,2026$ \\
General-capability test & Zero-shot MMLU after Order-1 \\
Mechanism sequence & DBpedia $\rightarrow$ Amazon \\
Mechanism trace & 625 Task-2 steps, 64 recorded points \\

LoRA rank / alpha & $r=8$, $\alpha=32$ \\
LoRA dropout & 0.1 \\
Target modules & \texttt{q\_proj} and \texttt{v\_proj} \\
Optimizer & AdamW, $\beta_1=0.9$, $\beta_2=0.999$, $\mathrm{lr}=10^{-4}$ \\
Schedule / precision & Constant schedule, bf16, weight decay 0 \\
Training batch & Micro-batch 1, gradient accumulation 8 \\
Input / evaluation & 512-token input; greedy decoding, 50 new tokens \\
BOD cohort soft orthogonality & $\lambda_{\mathrm{orth}}=0.5$ \\
BOD cohort effective-update weight & $\lambda_{\mathrm{BSR}}=1.0$ \\
Projector construction & $\tau=0.93$; $4\leq k_\ell\leq20$; wide cap 64; 128 token features per layer/task \\
\bottomrule
\end{tabular}
\end{table}

\subsection{Method and Evaluation Protocol}
\label{app:protocol}

All compared methods use the same backbone, target modules, LoRA rank, optimizer, and task-training protocol whenever their parameterization permits a direct match. Historical data are not replayed by the rehearsal-free methods. Replay uses a 500-example memory and a 5\% replay fraction. For every task order, final benchmark metrics are computed after the last task has been learned. Mechanism traces use the Task-1 checkpoint as the anchor and compare the same tracked LoRA layers across methods. Each reported trace statistic is aggregated across seeds before the endpoint values are reported.

The Standard CL runs use one training epoch per task and 3,625 optimizer steps across each four-task cell, with seeds $\{38,42,2026\}$. Each cell contains 29,000 training examples, and all 7,600 entries in each imported CL\_Benchmark test file are evaluated. The v3 loader checks this count against the asset manifest and uses the complete imported file without an additional evaluation subsample. The importer preserves the benchmark JSON files, source commit, and file hashes; the original-dataset sampling provenance is not established by this loader. The backbone is frozen and LoRA is applied to the query and value projections. For the BOD-aware runs, the collector retains at most 128 valid-token input-feature rows per layer per task, in encounter order during training forwards. These are not 128 independently sampled examples. At each task boundary, newly collected rows are appended to the historical feature store and the SVD-based core is recomputed; the resulting projector is fixed throughout the next task. The rows are detached to CPU and are not re-encoded using a single final checkpoint. Historical raw examples are not replayed, but historical feature storage grows with the number of tasks. The retained core uses the squared-singular-value retention threshold $\tau=0.93$ with $4\leq k_\ell\leq20$; the rank cap can prevent reaching that threshold. We do not report an observed retained-energy distribution. The wide BSR basis is capped at 64 directions. The mechanism trace uses the same LoRA configuration. Unless stated otherwise, reported means aggregate these three seeds; standard deviations use the sample estimator with $\mathrm{ddof}=1$.

For hyperparameter selection, tunable baselines were evaluated using four predeclared candidates on a deterministic 10\% training holdout with Qwen3-4B, Order-1, and seed 42. Candidate values were selected by validation average accuracy, and the selected configuration was frozen before any test-set evaluation. Methods with fixed published or predeclared settings were not searched. The candidate definitions and selection manifests accompany the code rather than being repeated for every baseline here. For the diagnostic BSR extension, $\lambda_{\mathrm{BSR}}\in\{0.01,0.1,1,10\}$ was the only searched component; validation selected $1$, while $\lambda_{\mathrm{orth}}=0.5$, projector width, and WRP were fixed. No test-set metric was used for configuration choice.

\paragraph{Parameter budgets and optimizer state.}
For a layer with input and output dimensions $d_{\mathrm{in}}$ and $d_{\mathrm{out}}$, one rank-$r$ block stores $r(d_{\mathrm{in}}+d_{\mathrm{out}})$ parameters. Projected-LoRA and SFOR each retain one such block. After Task~1, SFOR trains only the $rd_{\mathrm{in}}$ routing parameters. O-LoRA and hard protection each store $t$ blocks after task $t$, with the same historical-block freezing rule. Accordingly, capacity growth is matched within the cumulative comparison. Feature collection, subspace storage, and projection add overhead to hard protection; equal adapter capacity does not imply equal runtime or total memory. Cross-family scores describe different resource choices.

The runner constructs a new optimizer at each task boundary. Within a task, WRP snapshots the routing parameters before AdamW, then overwrites them with the projected displacement after the step. It does not project or reset Adam's first- and second-moment buffers. Projection therefore changes the optimization trajectory while enforcing the parameter constraint. All reported benchmark runs use zero weight decay; the decay example in the analysis describes a general feasibility issue rather than a source of the measured benchmark effects.

\subsection{Effective-Update Regularization}
\label{app:bsr_objective}

To test whether direct magnitude control adds value beyond feasibility, we also compute a wide historical basis $V_{\mathrm{wide}}^{(\ell,t-1)}$, containing up to 64 retained directions from tasks before $t$, and define the current-block response matrix
\begin{equation}
    E_{\mathrm{wide}}^{(\ell,t)}
    =s_\ell\left[
      B_t^{(\ell)}A_t^{(\ell)}
      -B_{t,0}^{(\ell)}A_{t,0}^{(\ell)}
      \right]V_{\mathrm{wide}}^{(\ell,t-1)}.
    \label{eq:bsr_drift}
\end{equation}
The optional Bilinear Stabilization Regularization (BSR) term is
\begin{equation}
    \mathcal{L}_{\mathrm{BSR}}^{(t)}
    =\frac{1}{|\mathcal{I}|}
      \sum_{\ell\in\mathcal{I}}
      \frac{\left\|E_{\mathrm{wide}}^{(\ell,t)}\right\|_F^2}
           {n_{\mathrm{el}}(E_{\mathrm{wide}}^{(\ell,t)})},
    \label{eq:bsr_loss}
\end{equation}
where $n_{\mathrm{el}}(E_{\mathrm{wide}}^{(\ell,t)})$ is the number of entries in the drift matrix. Unlike a penalty on either factor, Eq.~(\ref{eq:bsr_loss}) acts on their product. The full diagnostic control uses
\begin{equation}
    \mathcal{L}^{(t)}
    =\mathcal{L}_{\mathrm{task}}^{(t)}
     +\lambda_{\mathrm{orth}}\mathcal{L}_{\mathrm{orth}}^{(t)}
     +\lambda_{\mathrm{BSR}}\mathcal{L}_{\mathrm{BSR}}^{(t)}.
    \label{eq:olora_bsr_objective}
\end{equation}
for $t>1$ in addition to hard protection, where $\lambda_{\mathrm{orth}}$ and $\lambda_{\mathrm{BSR}}$ weight the two penalties. The ablation tests hard protection and this regularizer separately. The core-subspace guarantee already follows from hard protection. The wider-subspace regularizer shows no independent performance gain and is not part of the primary intervention.

\paragraph{O-LoRA controls.}
\label{app:bsr_protocol}

O-LoRA hard protection retains the native task-wise soft orthogonality objective and adds three operations to the current block: task-start retraction $A_{t,0}\leftarrow A_{t,0}P_{\mathrm{null}}^{(\ell,t-1)}$, current-block $A_t$ gradient projection, and post-step WRP. Historical O-LoRA blocks are neither projected nor updated. The diagnostic effective-update component computes $E_{\mathrm{wide}}^{(\ell,t)}$ in Eq.~(\ref{eq:bsr_drift}) on the wide historical basis and adds $\lambda_{\mathrm{BSR}}\mathcal{L}_{\mathrm{BSR}}$ to the task objective; the full BSR control combines this term with hard protection. For SFOR, the shared $B$ factor is frozen after Task~1, while the shared $A$ factor receives gradient projection and WRP without state retraction.

\label{app:metrics}

\subsection{Metric Definitions}
\label{app:cl_metrics}

Let $a_{t,i}$ denote the evaluation accuracy on task $i$ immediately after training task $t$, where $1\leq i\leq t\leq N$; $N=4$ in the main benchmark. We compute
\begin{equation}
    \mathrm{AA}=\frac{1}{N}\sum_{i=1}^{N}a_{N,i},
    \qquad
    \mathrm{LA}=\frac{1}{N}\sum_{i=1}^{N}a_{i,i},
    \label{eq:app_aa_la}
\end{equation}
\begin{equation}
    \mathrm{BWT}=\frac{1}{N-1}\sum_{i=1}^{N-1}
    \left(a_{N,i}-a_{i,i}\right),
    \label{eq:app_bwt}
\end{equation}
and
\begin{equation}
    \mathrm{FM}=\frac{1}{N-1}\sum_{i=1}^{N-1}
    \left(\max_{t\in\{i,\ldots,N-1\}}a_{t,i}-a_{N,i}\right).
    \label{eq:app_fm}
\end{equation}
AA and LA are reported as percentages; BWT and FM are percentage-point differences. BWT is closer to zero when less forgetting occurs, while lower signed FM indicates better retention. The maximum in Eq.~(\ref{eq:app_fm}) is taken from the checkpoint that first learns task $i$ through the checkpoint immediately before the final one. Consequently, FM may be negative when the final model improves beyond every preceding score on an old task.

\paragraph{Effective-update measurements.}
\label{app:mechanism_metrics}

For layer $\ell$ and optimizer step $q$, let $(A_{\ell,0},B_{\ell,0})$ be the Task-1 anchor and $(A_{\ell,q},B_{\ell,q})$ the current factors. We compute
\begin{equation}
    \Delta W_{\mathrm{eff}}^{(\ell)}(q)
    =s_{\ell}B_{\ell,q}A_{\ell,q}
    -s_{\ell}B_{\ell,0}A_{\ell,0}.
    \label{eq:app_delta_w}
\end{equation}
With $V_{\mathrm{core}}^{(\ell)}$ denoting the retained historical right-singular directions and $P_{\mathrm{null}}^{(\ell)}=I-V_{\mathrm{core}}^{(\ell)}\left(V_{\mathrm{core}}^{(\ell)}\right)^{\top}$, the cross-layer root-sum-of-squares metrics are
\begin{equation}
    D_{\mathrm{eff}}(q)
    =\left(\sum_{\ell\in\mathcal{I}}\|\Delta W_{\mathrm{eff}}^{(\ell)}(q)\|_F^2\right)^{1/2},
    \qquad
    D_{\mathrm{old}}(q)
    =\left(\sum_{\ell\in\mathcal{I}}\|\Delta W_{\mathrm{eff}}^{(\ell)}(q)V_{\mathrm{core}}^{(\ell)}\|_F^2\right)^{1/2},
    \label{eq:app_deff_dold}
\end{equation}
\begin{equation}
    D_{\mathrm{new}}(q)
    =\left(\sum_{\ell\in\mathcal{I}}\|\Delta W_{\mathrm{eff}}^{(\ell)}(q)P_{\mathrm{null}}^{(\ell)}\|_F^2\right)^{1/2}.
    \label{eq:app_dnew}
\end{equation}
For the cumulative O-LoRA trace, we also retain the raw current-block
quantity $D_{\mathrm{block}}$, the root-sum-of-squares norm of the change in
$sB_tA_t$ for the trainable block. It is reported separately from the
complementary-subspace quantity in Eq.~(\ref{eq:app_dnew}).
The normalized BOD residual share is
\begin{equation}
    \rho_{\mathrm{BOD}}(q)=
    \frac{D_{\mathrm{old}}(q)}{D_{\mathrm{eff}}(q)+\varepsilon},
    \label{eq:app_rho_bod}
\end{equation}
and is reported as a percentage in the endpoint mechanism tables. The diagnostics operate on composed updates, which are invariant under the factor rescaling $BA=(cB)(A/c)$ for any nonzero scalar $c$.

\clearpage
\section{Complete Results}
\subsection{Continual-Learning Results}
\begin{table}[H]
\centering
\small
\caption{Complete Standard CL metrics, including learning accuracy, with Qwen3-8B. Each row averages the three task orders and seeds $\{38,42,2026\}$. Order-wise AA appears in Table~\ref{tab:standard_cl_qwen8b}; all seed statistics are in Appendix~\ref{app:uncertainty}. AA and LA are percentages; BWT and signed FM are percentage points. Bold and underline denote the best and second-best values.}
\label{tab:complete_cl_metrics}
\begin{tabular}{lcccc}
\toprule
\textbf{Method} & \textbf{AA} & \textbf{LA} & \textbf{BWT} & \textbf{FM} \\
\midrule
\multicolumn{5}{l}{\emph{Three-order comparison}} \\
Replay & \underline{81.13} & \underline{82.10} & -1.30 & 1.39 \\
EWC & 80.87 & 82.04 & -1.56 & 1.59 \\
SFOR & 80.45 & 81.10 & \underline{-0.86} & \underline{0.91} \\
Projected-LoRA & 80.38 & \textbf{82.24} & -2.47 & 2.50 \\
O-LoRA & 80.27 & 81.87 & -2.13 & 2.20 \\
O-LoRA + hard protection & \textbf{81.30} & 81.59 & \textbf{-0.38} & \textbf{0.43} \\
IncLoRA & 80.24 & 82.06 & -2.43 & 2.49 \\
SD-LoRA & 80.19 & 82.05 & -2.48 & 2.53 \\
N-LoRA & 80.12 & 81.78 & -2.21 & 2.28 \\
LwF & 79.90 & 81.02 & -1.50 & 1.53 \\
Seq-LoRA & 79.68 & \textbf{82.24} & -3.42 & 3.47 \\
\bottomrule
\end{tabular}
\end{table}
\begin{table}[H]
\centering
\small
\caption{Final average accuracy on Qwen3-8B for each Standard CL order. Each value is averaged over seeds $\{38,42,2026\}$.}
\label{tab:standard_cl_qwen8b}
\begin{tabular}{lcccc}
\toprule
\textbf{Method} & \textbf{Order-1} & \textbf{Order-2} & \textbf{Order-3} & \textbf{Avg} \\
\midrule
Replay & \textbf{81.68} & \textbf{81.21} & 80.49 & \underline{81.13} \\
EWC & 80.94 & 80.87 & 80.80 & 80.87 \\
SFOR & 80.41 & 80.06 & 80.88 & 80.45 \\
Projected-LoRA & 80.44 & 80.47 & 80.24 & 80.38 \\
O-LoRA & 79.72 & 79.47 & \textbf{81.61} & 80.27 \\
O-LoRA + hard protection & 81.21 & 81.12 & 81.58 & \textbf{81.30} \\
IncLoRA & 80.17 & 79.16 & 81.41 & 80.24 \\
SD-LoRA & 79.67 & 79.54 & 81.36 & 80.19 \\
N-LoRA & 79.38 & 79.57 & 81.42 & 80.12 \\
LwF & 80.10 & 79.22 & 80.38 & 79.90 \\
Seq-LoRA & 80.30 & 78.86 & 79.88 & 79.68 \\
\bottomrule
\end{tabular}
\end{table}

\subsection{General-capability evaluation}
\label{app:mmlu}

We evaluate zero-shot MMLU on final Qwen3-8B checkpoints after Order-1, without training or hyperparameter tuning on MMLU. Values are means over seeds $\{38,42,2026\}$; $\Delta$ is the change from the base model mean. Standard deviations appear in Appendix~\ref{app:uncertainty}.

\begin{table}[H]
\centering
\small
\caption{General-capability retention measured by zero-shot MMLU.}
\label{tab:mmlu_qwen8b}
\begin{tabular}{lcc}
\toprule
\textbf{Method} & \textbf{MMLU Mean} & \textbf{$\Delta$ vs. Base} \\
\midrule
Qwen3-8B Base & 74.73 & 0.00 \\
SFOR & \textbf{74.71} & \textbf{-0.01} \\
Projected-LoRA & 74.34 & -0.38 \\
O-LoRA & 73.98 & -0.75 \\
O-LoRA + hard protection & 74.59 & -0.14 \\
Seq-LoRA & 73.88 & -0.85 \\
\bottomrule
\end{tabular}
\end{table}

SFOR has the smallest mean drop relative to the base model; hard protection remains within 0.14 percentage points of the base score.

\subsection{Mechanism Endpoints}
\label{app:mechanism_results}

Tables~\ref{tab:sfor_family_mechanism} and~\ref{tab:olora_family_mechanism} report all component endpoints from the trajectories in Figure~\ref{fig:main_bod_dynamics}. Standard deviations are collected in Appendix~\ref{app:uncertainty}.

\begin{table}[H]
\centering
\small
\caption{Endpoint mechanism values for the Projected-LoRA family. $\rho_{\mathrm{BOD}}$ is reported in percentages; the effective-update norms use native units.}
\label{tab:sfor_family_mechanism}
\begin{tabular}{lcccc}
\toprule
\textbf{Method} & $D_{\mathrm{eff}}$ & $D_{\mathrm{old}}$ & $\rho_{\mathrm{BOD}}$ & $D_{\mathrm{new}}$ \\
\midrule
Projected-LoRA & 7.540 & 1.443 & 19.12 & 7.401 \\
Projected-LoRA + WRP & 7.466 & 1.227 & 16.44 & 7.364 \\
Projected-LoRA + Freeze-$B$ & 4.538 & 0.697 & 15.33 & 4.484 \\
SFOR & 4.293 & $2.08\times10^{-4}$ & 0.005 & 4.293 \\
\bottomrule
\end{tabular}
\end{table}

\begin{table}[H]
\centering
\small
\caption{Endpoint mechanism values for the O-LoRA component family. $\rho_{\mathrm{BOD}}$ is reported in percentages; the effective-update norms use native units. Here $D_{\mathrm{block}}$ is the effective drift of the trainable current block, whereas $D_{\mathrm{old}}$ is its response on the retained historical basis.}
\label{tab:olora_family_mechanism}
\scriptsize
\begin{tabular}{lcccc}
\toprule
\textbf{Method} & $D_{\mathrm{eff}}$ & $D_{\mathrm{old}}$ & $\rho_{\mathrm{BOD}}$ & $D_{\mathrm{block}}$ \\
\midrule
O-LoRA & 6.712 & 0.518 & 7.72 & 6.712 \\
O-LoRA + hard protection & 7.227 & $1.10\times10^{-4}$ & 0.002 & 7.227 \\
O-LoRA + effective-update regularization & 6.639 & 0.511 & 7.69 & 6.639 \\
O-LoRA + BSR & 7.303 & $1.19\times10^{-4}$ & 0.002 & 7.303 \\
\bottomrule
\end{tabular}
\end{table}

\subsection{Component Results}

\label{app:ablation_details}

The following tables give the completed component cohorts summarized in the main text. Every row averages seeds $\{38,42,2026\}$. The O-LoRA component table reports the nested intervention used for the mechanism analysis. Corresponding standard deviations are listed in Appendix~\ref{app:uncertainty}.

\begin{table}[H]
\centering
\small
\caption{Projected-LoRA component cohorts used in the shared-adapter mechanism analysis. Mechanism attribution uses the trace in Appendix~\ref{app:mechanism_results}.}
\label{tab:projected_component_ablation}
\begin{tabular}{lcccc}
\toprule
\textbf{Method} & \textbf{AA} & \textbf{LA} & \textbf{BWT} & \textbf{FM} \\
\midrule
Projected-LoRA & 80.38 & 82.24 & -2.47 & 2.50 \\
Projected-LoRA + WRP & 80.474 & 82.190 & -2.288 & 2.292 \\
Projected-LoRA + Freeze-$B$ & 80.387 & 80.891 & -0.673 & 0.680 \\
SFOR (Freeze-$B$ + WRP) & 80.45 & 81.10 & -0.86 & 0.91 \\
\bottomrule
\end{tabular}
\end{table}

\begin{table}[H]
\centering
\small
\caption{O-LoRA component ablation on Order-1. Hard protection denotes task-start retraction, current-block $A_t$ gradient projection, and post-step WRP. Effective-update regularization is the BSR loss without hard protection; the full BSR control combines both components.}
\label{tab:olora_bsr_ablation}
\begin{tabular}{lcccc}
\toprule
\textbf{Method} & \textbf{AA} & \textbf{LA} & \textbf{BWT} & \textbf{FM} \\
\midrule
O-LoRA (matched rerun) & 79.78 & 81.83 & -2.73 & 2.74 \\
O-LoRA + initialization retraction & 80.81 & 81.71 & -1.20 & 1.21 \\
O-LoRA + initialization retraction + $A$-gradient projection & 81.33 & 81.60 & -0.36 & 0.36 \\
O-LoRA + hard protection & 81.21 & 81.55 & -0.46 & 0.46 \\
O-LoRA + effective-update regularization & 79.95 & 81.85 & -2.53 & 2.53 \\
O-LoRA + BSR & 81.15 & 81.59 & -0.58 & 0.59 \\
\bottomrule
\end{tabular}
\end{table}

\clearpage
\section{Standard Deviations}
\label{app:uncertainty}
All standard-deviation tables are collected here. Each uses seeds $\{38,42,2026\}$ and the sample estimator ($\mathrm{ddof}=1$, $n=3$); task-order coverage is stated in the captions. Accuracy, BWT, and FM standard deviations are in percentage points. Results without complete seed-level exports are reported as means only.
\begin{table}[H]
\centering
\scriptsize
\caption{Available seed standard deviations for the four-task experiments, including hard protection and component controls. Each row uses three seeds within its stated cohort ($n=3$); all metric entries are percentage points. Matched reruns correspond to the component results in Table~\ref{tab:olora_bsr_ablation}.}
\label{tab:cl_seed_std}
\begin{tabular}{p{0.48\linewidth}ccccc}
\toprule
\textbf{Method} & \textbf{Order} & \textbf{AA std} & \textbf{LA std} & \textbf{BWT std} & \textbf{FM std} \\
\midrule
O-LoRA (matched rerun) & 1 & 0.66 & 0.43 & 0.50 & 0.50 \\
O-LoRA + hard protection & 1 & 0.40 & 0.36 & 0.11 & 0.11 \\
O-LoRA + hard protection & 2 & 0.60 & 0.25 & 0.47 & 0.46 \\
O-LoRA + hard protection & 3 & 0.26 & 0.21 & 0.58 & 0.60 \\
Projected-LoRA + Freeze-$B$ & 1 & 0.07 & 0.12 & 0.06 & 0.06 \\
Projected-LoRA + WRP & 1 & 0.64 & 0.28 & 0.96 & 0.97 \\
O-LoRA + initialization retraction & 1 & 0.41 & 0.30 & 0.24 & 0.23 \\
O-LoRA + initialization retraction + $A$-gradient projection & 1 & 0.33 & 0.32 & 0.11 & 0.12 \\
O-LoRA + effective-update regularization & 1 & 0.70 & 0.44 & 0.52 & 0.52 \\
O-LoRA + BSR (matched rerun) & 1 & 0.33 & 0.25 & 0.12 & 0.11 \\
\bottomrule
\end{tabular}
\end{table}
\begin{table}[H]
\centering
\small
\caption{Seed standard deviations for zero-shot MMLU (percentage points).}
\label{tab:mmlu_seed_std}
\begin{tabular}{lc}
\toprule
\textbf{Method} & \textbf{MMLU Std} \\
\midrule
Qwen3-8B Base & 0.00 \\
SFOR & 0.10 \\
Projected-LoRA & 0.30 \\
O-LoRA & 0.03 \\
O-LoRA + hard protection & 0.10 \\
Seq-LoRA & 0.29 \\
\bottomrule
\end{tabular}
\end{table}

\begin{table}[H]
\centering
\scriptsize
\caption{Seed standard deviations for the Order-1 cross-backbone check (percentage points; $n=3$).}
\label{tab:cross_backbone_std}
\begin{tabular}{lccc}
\toprule
\textbf{Method} & \textbf{AA std} & \textbf{BWT std} & \textbf{FM std} \\
\midrule
Seq-LoRA & 3.78 & 4.14 & 4.09 \\
O-LoRA & 0.47 & 0.46 & 0.49 \\
O-LoRA + hard protection & 0.42 & 0.75 & 0.75 \\
Projected-LoRA & 1.32 & 1.23 & 1.21 \\
SFOR & 0.38 & 0.88 & 0.89 \\
\bottomrule
\end{tabular}
\end{table}

\begin{table}[H]
\centering
\scriptsize
\caption{Per-seed final metrics for the Mistral-7B-v0.3 Order-1 check. Values are percentages or percentage points.}
\label{tab:cross_backbone_seed}
\begin{tabular}{llccc}
\toprule
\textbf{Method} & \textbf{Seed} & \textbf{AA} & \textbf{BWT} & \textbf{FM} \\
\midrule
Seq-LoRA & 38 & 67.07 & $-18.88$ & 18.94 \\
Seq-LoRA & 42 & 69.66 & $-15.79$ & 15.79 \\
Seq-LoRA & 2026 & 74.52 & $-10.68$ & 10.83 \\
O-LoRA & 38 & 78.83 & $-4.66$ & 4.70 \\
O-LoRA & 42 & 79.29 & $-5.05$ & 5.13 \\
O-LoRA & 2026 & 79.78 & $-4.14$ & 4.14 \\
O-LoRA + hard protection & 38 & 81.42 & $-1.59$ & 1.61 \\
O-LoRA + hard protection & 42 & 80.97 & $-2.95$ & 2.95 \\
O-LoRA + hard protection & 2026 & 80.59 & $-2.82$ & 2.85 \\
Projected-LoRA & 38 & 74.23 & $-10.21$ & 10.21 \\
Projected-LoRA & 42 & 74.32 & $-10.72$ & 10.73 \\
Projected-LoRA & 2026 & 76.56 & $-8.38$ & 8.42 \\
SFOR & 38 & 81.48 & $-1.28$ & 1.28 \\
SFOR & 42 & 81.42 & $-1.93$ & 2.00 \\
SFOR & 2026 & 82.10 & $-0.19$ & 0.23 \\
\bottomrule
\end{tabular}
\end{table}

\begin{table}[H]
\centering
\small
\caption{Paired Mistral-7B-v0.3 differences (intervention minus baseline), in percentage points. Positive AA/BWT and negative FM favor the intervention. Each row compares the same seed and Order-1; these three pairs describe the observed cohort.}
\label{tab:mistral_paired}
\begin{tabular}{llrrr}
\toprule
Comparison & Seed & $\Delta$AA & $\Delta$BWT & $\Delta$FM \\
\midrule
SFOR $-$ Projected-LoRA & 38 & +7.24 & +8.93 & -8.93 \\
SFOR $-$ Projected-LoRA & 42 & +7.11 & +8.79 & -8.74 \\
SFOR $-$ Projected-LoRA & 2026 & +5.54 & +8.19 & -8.18 \\
Hard protection $-$ O-LoRA & 38 & +2.59 & +3.07 & -3.10 \\
Hard protection $-$ O-LoRA & 42 & +1.68 & +2.10 & -2.18 \\
Hard protection $-$ O-LoRA & 2026 & +0.81 & +1.32 & -1.29 \\
\bottomrule
\end{tabular}
\end{table}

The official exact-match evaluator records method- and seed-dependent illegal generated labels after later tasks, most prominently for Seq-LoRA and Projected-LoRA on DBpedia. These outputs are counted as incorrect under the frozen label contract; the reported metrics use the complete 7,600-example test files and the same parser for every method. SFOR and hard protection show substantially lower illegal-label rates in this cohort.

\begin{table}[H]
\centering
\scriptsize
\caption{Endpoint standard deviations for the Projected-LoRA mechanism trace ($n=3$). Effective-update norms use native units; $\rho_{\mathrm{BOD}}$ is reported in percentage points.}
\label{tab:projected_mechanism_std}
\begin{tabular}{lccc}
\toprule
\textbf{Method} & $D_{\mathrm{eff}}$ std & $D_{\mathrm{old}}$ std & $\rho_{\mathrm{BOD}}$ std \\
\midrule
Projected-LoRA & 0.1667 & 0.1156 & 1.10 \\
Projected-LoRA + WRP & 0.2595 & 0.0397 & 0.17 \\
Projected-LoRA + Freeze-$B$ & 0.2273 & 0.0844 & 1.36 \\
SFOR & 0.2164 & 0.0000069 & 0.00028 \\
\bottomrule
\end{tabular}
\end{table}

\begin{table}[H]
\centering
\scriptsize
\caption{Endpoint standard deviations for WRP constraint diagnostics in the Projected-LoRA trace ($n=3$). Pre-violation is scaled by $10^{-3}$; post-violation is scaled by $10^{-9}$.}
\label{tab:projected_output_std}
\begin{tabular}{lcc}
\toprule
\textbf{Method} & \textbf{WRP pre-violation std} & \textbf{WRP post-violation std} \\
\midrule
Projected-LoRA & 0.0000 & 0.0000 \\
Projected-LoRA + WRP & 1.0920 & 0.1628 \\
Projected-LoRA + Freeze-$B$ & 0.0000 & 0.0000 \\
SFOR & 0.3001 & 0.2474 \\
\bottomrule
\end{tabular}
\end{table}

\begin{table}[H]
\centering
\scriptsize
\caption{Endpoint standard deviations for the O-LoRA component trace ($n=3$). Effective-update norms use native units; $\rho_{\mathrm{BOD}}$ is reported in percentage points.}
\label{tab:olora_mechanism_std}
\resizebox{\textwidth}{!}{%
\begin{tabular}{lccc}
\toprule
\textbf{Method} & $D_{\mathrm{eff}}$ std & $D_{\mathrm{old}}$ std & $\rho_{\mathrm{BOD}}$ std \\
\midrule
O-LoRA & 0.1577 & 0.013210 & 0.0552 \\
O-LoRA + hard protection & 0.2454 & 0.000014 & 0.00015 \\
O-LoRA + effective-update regularization & 0.2333 & 0.017783 & 0.0480 \\
O-LoRA + BSR & 0.1285 & 0.000016 & 0.00024 \\
\bottomrule
\end{tabular}
}
\end{table}

\begin{table}[H]
\centering
\scriptsize
\caption{Endpoint standard deviations for WRP diagnostics in the O-LoRA component trace ($n=3$). Pre-violation is scaled by $10^{-3}$ and post-violation by $10^{-12}$.}
\label{tab:olora_wrp_std}
\begin{tabular}{lcc}
\toprule
\textbf{Method} & \textbf{WRP pre-violation std} & \textbf{WRP post-violation std} \\
\midrule
O-LoRA & 0.0000 & 0.0000 \\
O-LoRA + hard protection & 0.0211 & 4.7308 \\
O-LoRA + effective-update regularization & 0.0000 & 0.0000 \\
O-LoRA + BSR & 0.0160 & 1.2622 \\
\bottomrule
\end{tabular}
\end{table}

\clearpage
\end{document}